\documentclass[runningheads]{llncs}
\usepackage[T1]{fontenc}

\usepackage{hyperref}
\usepackage{amsmath}
\usepackage{caption}
\usepackage{array}
\usepackage{subcaption}
\usepackage{booktabs}
\usepackage[parfill]{parskip}

\usepackage{graphicx}
\begin{document}

\title{Fairness measures for \\biometric quality assessment}

\author{André Dörsch\inst{1} \and
Torsten Schlett\inst{1} \and
Peter Munch\inst{2} \and
Christian Rathgeb\inst{1} \and
Christoph Busch\inst{1}}

\authorrunning{A. Dörsch et al.}

\institute{
da/sec – Biometrics and Security Research Group\\ Hochschule Darmstadt, Germany\\
\email{firstname.lastname@h-da.de}\\
\url{https://dasec.h-da.de/} \and
Department of Applied Mathematics and Computer Science\\ Technical University of Denmark, Denmark\\
\email{s183718@dtu.dk}\\
\url{https://www.compute.dtu.dk/}}

\maketitle             

\begin{abstract}
Quality assessment algorithms measure the quality of a captured biometric sample.
Since the sample quality strongly affects the recognition performance of a biometric system, it is essential to only process samples of sufficient quality and discard samples of low-quality. Even though quality assessment algorithms are not intended to yield very different quality scores across demographic groups, quality score discrepancies are possible, resulting in different discard ratios. To ensure that quality assessment algorithms do not take demographic characteristics into account when assessing sample quality and consequently to ensure that the quality algorithms perform equally for all individuals, it is crucial to develop a fairness measure. In this work we propose and compare multiple fairness measures for evaluating quality components across demographic groups. Proposed measures, could be used as potential candidates for an upcoming standard in this important field.  

\keywords{Biometrics \and Fairness \and Quality \and Quality component \and Quality measure \and Demographic differentials \and Quality assessment.}
\end{abstract}

\section{Introduction}

\begin{figure}[htb]
\centering
    \includegraphics[width=\textwidth]{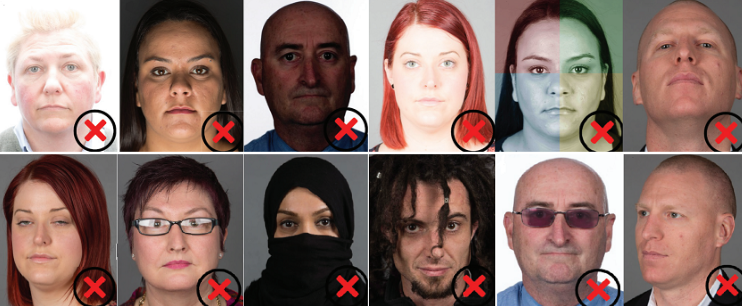}
    \caption{Various examples of face image defects (i.e. factors) of a captured sample that negatively impact the recognition performance. As a result, the images shown are not compliant with requirements formulated in ISO/IEC 39794-5 \cite{ISO-IEC-39794-5-G3-FaceImage-191015}. Facial images taken from \cite{ISO-IEC-39794-5-G3-FaceImage-191015}.}
    \label{quality_fiq}
\end{figure}

Biometric systems are widely used due to their reliable  recognition performance. They are an integral part of ``automated decision systems'' which support applications such as border controls, law enforcement, physical access control and forensic investigations \cite{ISO-IEC-24741-TR-BiometricTutorial}. For a biometric system to function accurately, the captured biometric sample \cite{ISO-IEC-2382-37-220330} (e.g. a fingerprint or a facial image) must be of high quality. However, different factors may negatively impact biometric quality such as varying environmental conditions \cite{JainFlynnRoss-HandbookOfBiometrics-Springer-2007}. Hence, capturing high-quality biometric samples still remains a difficult task. Examples of factors that have a negative impact on the quality of a face image can be seen in Figure \ref{quality_fiq}. To this end, great efforts have been placed into developing quality assessment algorithms for various biometric characteristics to estimate the quality of a captured biometric sample and ensure that its quality is sufficient.

In the context of fingerprint recognition, the National Institute of Standards and Technology (NIST) introduced the first open-source and publicly available fingerprint quality assessment algorithm, NIST Fingerprint Image Quality (NFIQ), in 2004 \cite{Tabassi-NFIQ1-NISTIR-7151-2004}. The current NFIQ 2 \cite{NIST-NFIQ2-FingerprintImageQuality-2021} represents a refined version and has proven to be the de facto standard in this area. 
By dividing the quality assessment into different quality components, actionable feedback as well as a certain degree of explainability of the biometric sample quality are made possible. Inspired by this framework approach, the German Federal Office for Information Security recently introduced a novel Face Image Quality Assessment (FIQA) scheme named ``Open Source Face Image Quality'' (OFIQ)\footnote{\url{https://github.com/BSI-OFIQ/OFIQ-Project}}.

In most cases, the result of a biometric quality assessment is a quality score \cite{ISO-IEC-29794-5-DIS-FaceQuality-240129}, \cite{ISO-IEC-29794-1-QualityFramework-090205} (QS), which is a single scalar value, representing the captured biometric sample quality. Alternatively, the output of a biometric quality assessment can also be a vector of quality values (or components), measuring various quality-related properties \cite{Schlett-FIQA-LiteratureSurvey-CSUR-2021}. While there already exists numerous FIQA methods (see e.g. \cite{ou2021sdd}, \cite{boutros2023cr}, \cite{Meng-FRwithFQA-MagFace-CVPR-2021}, \cite{kolf2024grafiqsfaceimagequality}), the OFIQ algorithm is expected to be most influential, since it is used as a reference implementation for the international standard ISO/IEC DIS 29794-5 \cite{ISO-IEC-29794-5-DIS-FaceQuality-240129}.

Despite the ongoing improvement of biometric systems, topics as demographic fairness and potential biases in biometric systems and underlying algorithms have been hotly debated in recent years \cite{Drozdowski-BiasSurvey-TTS-2020}. Although biometric systems are not inherently designed to be discriminatory against any particular group, demographic performance differences cannot be entirely ruled out \cite{Rathgeb-FairnessExperts-TSM-2022}. According to Drozdowski et al. \cite{Drozdowski-BiasSurvey-TTS-2020}, the predominant concerns regarding bias and fairness have arisen primarily with face recognition systems. In addition, the aforementioned efforts towards establishing quality assessment algorithms for different biometric characteristics has shifted some research attention to fairness analysis of biometric sample quality assessment, e.g. in \cite{Terhorst-FQA-DemograhicBias-IJCB-2020}. 

To measure demographic performance differences between various demographic groups, ISO/IEC FDIS 19795-10 \cite{ISO-IEC-FDIS-19795-10-240701} introduced the term \textit{differential performance measure} (DPM), which is equivalent to the term \textit{demographic differential} listed in the standard ISO/IEC 2382-37 \cite{ISO-IEC-2382-37-220330}. In our context, a DPM is defined by a formula or algorithm that receives as input quality scores of different demographic groups and reflects how fair the underlying quality assessment algorithm is. Although ISO/IEC FDIS 19795-10 \cite{ISO-IEC-FDIS-19795-10-240701} specifies methods and statistical techniques for calculating DPMs, there is no dedicated standardised approach for assessing fairness of quality components across demographic groups. 

This research paper introduces and compares new statistical approaches based on quality score distributions, for assessing fairness of quality components across demographic groups. Proposed measures, could be used as potential candidates for defining a fairness measure in an upcoming standard. 
\section{Background and Related Work}

To ensure that quality algorithms provide equivalent results across demographic groups and investigate potential biases, various reports have been proposed in the scientific literature. 

In the current NIST FATE SIDD report \cite{1214921}, FIQA algorithms for five quality measures are evaluated to quantify demographic performance differentials. These performance differentials were investigated across six demographic groups. It was found that only certain algorithms for the quality measures \textit{Eyes Open 2} and \textit{Resolution} exhibit demographic bias, while several algorithms for the quality measures \textit{Mouth Open 2}, \textit{Underexposure} and \textit {Overexposure} exhibit demographic bias. As there is no standardised DPM for FIQA quality components yet, results shown were only visualized in the form of violin plots. In \cite{9909867} Babnik et al. investigated demographic biases in FIQA methods. Although no specific quality components were analysed, it was found that FIQA methods generally exhibit significant bias and tend to favour white individuals. Terhörst et al. \cite{Terhorst-FQA-DemograhicBias-IJCB-2020} evaluated FIQA algorithms with respect to potential bias in race and age. It was found, that for all evaluated FIQA algorithms, demographic performance differentials were observed.

These reports and studies demonstrate the importance of developing a standardised method for measuring demographic performance differentials in quality assessments and underlying algorithms in order to reveal potential biases in quality components.

\section{Differential Performance Measures}
\subsection{Gini Coefficient}
\label{sec:gini}
The Gini coefficient (GC) is a statistical measure of dispersion of a set of numbers \cite{1912vamu}. This index, originally used to calculate income inequality, can be applied to various scenarios, including biometric measures. One biometric DPM based on the GC can be found, for example, in the ISO/IEC FDIS 19795-10 \cite{ISO-IEC-FDIS-19795-10-240701} standard for calculating performance differences for multiple groups. Since there is not yet a standardised DPM for assessing the fairness of quality assessment across demographic groups, we decided to use the GC as the backbone for this approach. Therefore, either the mean or median quality scores for any quality component \textit{Q} across each demographic group \textit{$d_i$} are utilized as inputs to the GC as follows:

\begin{equation}
\label{eq:gini-coefficient}
\text{GC} = \left( \frac{n}{n-1} \right) \left( \frac{\sum_{i}^{n} \sum_{j}^{n} \left| Q_{d_i} - Q_{d_j} \right|}{2 n^2 \overline{Q}} \right) \quad \forall d_i, d_j \in D
\end{equation}

where \( n \) represents the number of demographic groups, \( Q_{d_i} \) is either the mean or median quality score of the demographic group \(d_i\) and  \( D \) is the set of all demographic groups to be evaluated. However, a disadvantage of using median quality scores over mean quality scores is that, given slightly different demographic distributions of quality scores with relatively few outliers, all groups may receive exactly the same median score, even if there exists a slight bias. In addition, as previously for other DPMs listed in ISO/IEC FDIS 19795-10 \cite{ISO-IEC-FDIS-19795-10-240701}, we adopt the approach of Howard et al. \cite{Howard-GARBE-ICPR-2022} by multiplying our result by a factor of $n/(n-1)$
to account for group self-comparisons ($i=j$ in equation \ref{eq:gini-coefficient}), which can be especially relevant for smaller group numbers ($n$).

\begin{figure}[htb]
\centering
\begin{minipage}{0.45\textwidth}
    \centering
    \includegraphics[width=\textwidth]{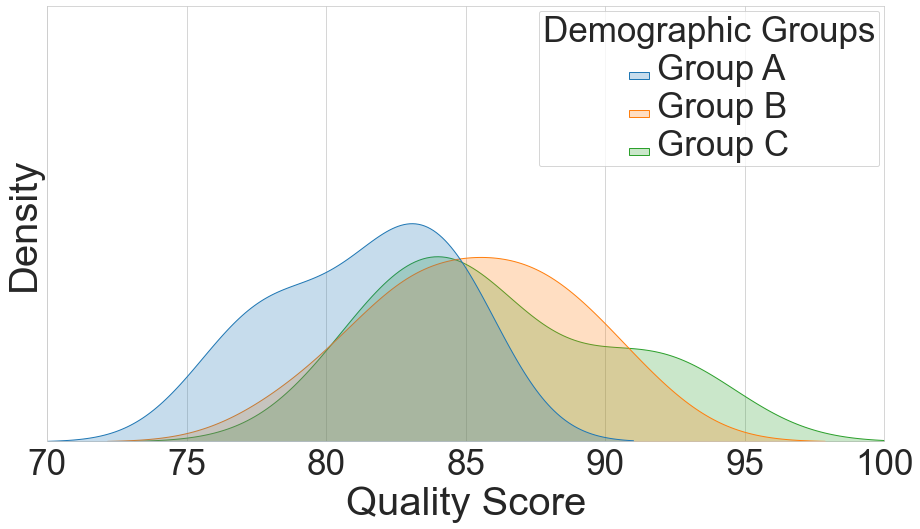}
    \caption{Fictitious quality component \textit{$Q_1$} (Slightly biased): KDE Plot of the demographic score distribution} \label{fig1}
\end{minipage}\hfill
\begin{minipage}{0.45\textwidth}
    \centering
    \captionof{table}{Fictitious quality component \textit{$Q_1$} (Slightly biased): Synthetic Mean and Median Quality Scores of different demographic groups}\label{tab1}
    \begin{tabular}{|l|c|c|c|}
    \hline
    & Group A & Group B & Group C \\
    \hline
    Mean & 81.3 & 85.3 & 86.1 \\
    \hline
    Median & 82 & 85.5 & 85 \\
    \hline
    \end{tabular}
\end{minipage}
\end{figure}

\begin{figure}[htb]
\centering
\begin{minipage}{0.45\textwidth}
    \centering
    \includegraphics[width=\textwidth]{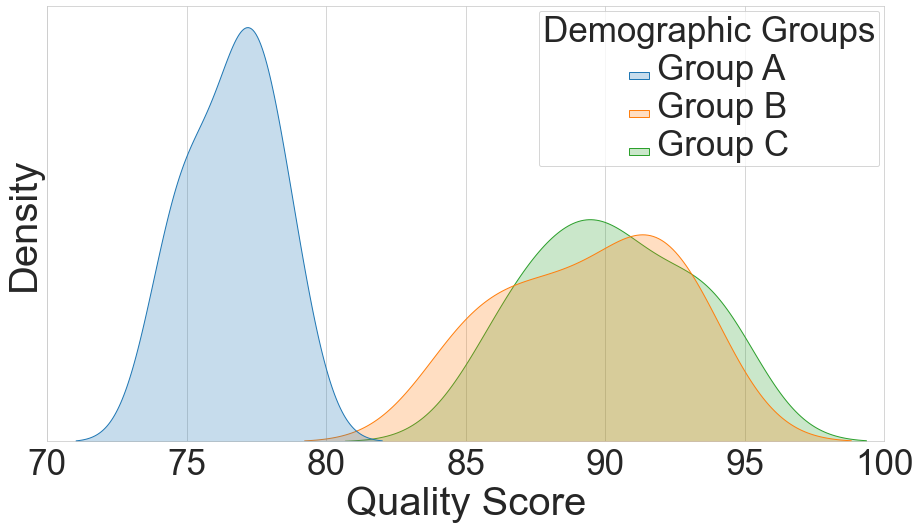}
    \caption{Fictitious quality component \textit{$Q_2$} (Strongly biased): KDE Plot of the demographic score distribution} \label{fig2}
\end{minipage}\hfill
\begin{minipage}{0.45\textwidth}
    \centering
      \captionof{table}{Fictitious quality component \textit{$Q_2$} (Strongly biased): Synthetic Mean and Median Quality Scores of different demographic groups}\label{tab2}
    \begin{tabular}{|l|c|c|c|}
    \hline
    & Group A & Group B & Group C \\
    \hline
    Mean & 76.6 & 89.4 & 90.2 \\
    \hline
    Median & 77 & 90 & 90 \\
    \hline
    \end{tabular}
\end{minipage}
\end{figure}

While the GC is used in many scenarios, one notable drawback is that it is rather insensitive to outliers. Table \ref{tab1} shows synthetically generated mean and median quality scores for three demographic groups. These quality scores can be interpreted as descriptive results of an arbitrary Quality Assessment Algorithm referring to any quality component \textit{Q}. The quality scores for the fictitious quality component \textit{$Q_1$} were generated in such a way that one of the three groups exhibits a slight bias (a deviation of approximately 4 to 5 quality score points on average) compared to the other two groups, as shown visually in Figure \ref{fig1}. Table \ref{tab2} shows another set of synthetically generated mean and median quality scores for the same three demographic groups. In this second fictitious quality component \textit{$Q_2$}, a more prominent bias is simulated (a deviation of approximately 13 to 14 quality score points on average) compared to the others, which is shown in Figure \ref{fig2}. The \textit{Sample Quality Fairness Rate} (SQFR), which follows a ``higher is better'' semantic outputs a fairness score in the range 0-1 and serves as our DPM. 

In this paper, the term SQFR for the general concept is annotated with a prefix depending on the method used. When using mean quality scores as input to the GC as fairness metric, the Mean-GC-SQFR is calculated as follows:
\begin{equation}
\text{Mean-GC-SQFR} = 1 - \text{GC}(Q_{d_n})   
\end{equation}

where $Q_{d_n}$ represents the mean quality scores of a quality component \(Q\) for a set of demographic groups \(D\).

When instead using median quality scores as input to the GC as fairness metric, the Median-GC-SQFR is calculated as follows:

\begin{equation}
\text{Median-GC-SQFR} = 1 - \text{GC}(Q_{d_n})   
\end{equation}
where $Q_{d_n}$ represents the median quality scores of a quality component \(Q\) for a set of demographic groups \(D\).

The SQFR scores for the setups discussed are shown in Table \ref{tab3}. For the two fictitious scenarios presented, the resulting SQFR scores are surprisingly high. This should not be the case, especially for the strongly biased quality component \textit{$Q_2$} (see e.g. Table \ref{tab2} or Figure \ref{fig2}), as a significant deviation should result in a general lower SQFR score. Even for quality component \textit{$Q_1$}, where one group slightly deviates from the others, one would not expect Mean-GC-SQFR or Median-GC-SQFR scores of \textit{0.98} and \textit{0.99}, which describe a near maximum fair system. To this end, to obtain a more reliable SQFR, the GC must be adjusted or alternative solutions developed as it does not sufficiently capture the underlying bias.

\begin{table}[htb]
\centering
\caption{SQFR results for quality component \textit{$Q_1$} and \textit{$Q_2$}}
\begin{tabular}{|c|c|c|}
\hline
 SQFR& Quality component \textit{$Q_1$} & Quality component \textit{$Q_2$} \\
\hline
Mean-GC-SQFR & 0.98 & 0.95 \\
\hline
Median-GC-SQFR & 0.99 & 0.95 \\
\hline
\end{tabular}
\vspace{0.5cm}
\label{tab3}
\end{table}

\subsection{Cubed Sample Quality Fairness Rate}
\label{sec:cgc}
To address the identified drawbacks of the traditional GC in Section \ref{sec:gini}, we provide an adapted approach, called Cubed Sample Quality Fairness Rate (CSQFR), which is designed to achieve lower fairness scores in scenarios with demographic bias. This approach places greater emphasis on biased scenarios by cubing the result, making cases of lower fairness more visible. The adapted CSQFR when using mean quality scores as input to the GC, resulting in the Mean-GC-CSQFR is calculated as follows:
\begin{equation}
 \text{Mean-GC-CSQFR} = (1 - \text{GC}(Q_{d_n}))^3   
\end{equation}
where $Q_{d_n}$ represents the mean quality scores\footnote{Due to the potential disadvantages of median scores compared to mean quality scores described in section \ref{sec:gini}, median values should not be used.} of a quality component \(Q\) for a set of demographic groups \(D\). Different Quality Score scenarios for three demographic groups are provided in Table \ref{tab4}.

\begin{table}[htb]
\centering
\caption{Comparison of Mean-GC-SQFR and Mean-GC-CSQFR scores for different scenarios}
\resizebox{1\textwidth}{!}{
\begin{tabular}{|l|c|c|c|c|c|c|c|}
\hline
Quality Score Scenarios & \begin{tabular}[c]{@{}c@{}}Mean QS\\ Group A\end{tabular} & \begin{tabular}[c]{@{}c@{}}Mean QS\\ Group B\end{tabular} & \begin{tabular}[c]{@{}c@{}}Mean QS\\ Group C\end{tabular} & \begin{tabular}[c]{@{}c@{}}Mean-GC-SQFR\end{tabular} & \begin{tabular}[c]{@{}c@{}}Mean-GC-CSQFR\end{tabular} \\ \hline
One group exhibits strong bias& 35& 95& 89& 0.73& 0.38 \\ \hline
One group exhibits slight bias& 67& 82& 89& 0.91&  0.75 \\ \hline
All groups receive different QS & 30& 50& 95& 0.63& 0.25 \\ \hline
All groups receive similar QS  & 84& 89& 87& 0.98& 0.94 \\ \hline
\end{tabular}
}
\label{tab4}
\end{table}

For the scenario \textit{One group exhibits strong bias} Group A received significantly lower quality scores on average than the other two groups. The Mean-GC-SQFR score assesses this scenario a score of 0.73, indicating moderate fairness. However, the Mean-GC-CSQFR approach more reliably reflects the underlying bias by assigning a value of 0.38 to this scenario. For the scenario \textit{One group exhibits slight bias} Group A received slightly lower quality scores on average than the other two groups. While the GC approach assesses this scenario with a high Mean-GC-SQFR score of 0.91, the CSQFR captures the underlying slight bias much better, resulting in a moderate Mean-GC-CSQFR fairness score of 0.74. Furthermore, for the last two scenarios \textit{All groups received different} and \textit{All groups received similar QS}, the Mean-GC-CSQFR reflects the average quality scores more precise than the traditional Mean-GC-SQFR.

\subsection{Low-Weighted-Mean Scores}
\label{sec:lbs}
Another DPM as an alternative to inputting the mean or medium quality scores into the GC, is our proposed approach of Low-Weighted-Mean (LWM) Scores. This method performs a linear weighting (from lowest to highest) of quality scores in a given demographic distribution, resulting in lower quality scores being weighted higher. Since captured biometric samples associated with lower scores would be rejected by more potential thresholds and this could disadvantage a group with in general lower quality scores more easily, this approach attempts to place a greater focus on fairness. This LWM weighting approach is calculated as follows:
For each quality score \textit{$q$}, a weight \textit{w} is calculated as follows:
\begin{align}
w(q) = 1 - \left(\frac{q - min(Q)}{max(Q) - min(Q)}\right)
\end{align} where \textit{Q} represents the union set of all quality scores across the demographic groups to be evaluated. This inverted min-max normalization ensures that our proposed method generalizes to quality scores at arbitrary scale while assigning higher weight to lower quality scores. For each quality score \textit{$q$} the calculated weights (multiple occurrences of the same quality score) are accumulated and used to calculate a weighted arithmetic mean of the corresponding quality scores. In the unlikely special case where $min(Q)=max(Q)$ (i.e. there exists only a single quality score across demographic groups), this single quality score could then alternatively be used as output.

The adapted SQFR when using the LWM, resulting in the LWM-GC-SQFR is calculated as follows:
\begin{equation}
 \text{LWM-GC-SQFR} = (1 - \text{GC}(\text{LWM}(Q_{d_n})))   
\end{equation}
where $Q_{d_n}$ represents the quality scores of a quality component \(Q\) for a set of demographic groups \(D\).

\begin{figure}[htb]
\centering
\begin{minipage}{0.45\textwidth}
    \centering
    \includegraphics[width=\textwidth]{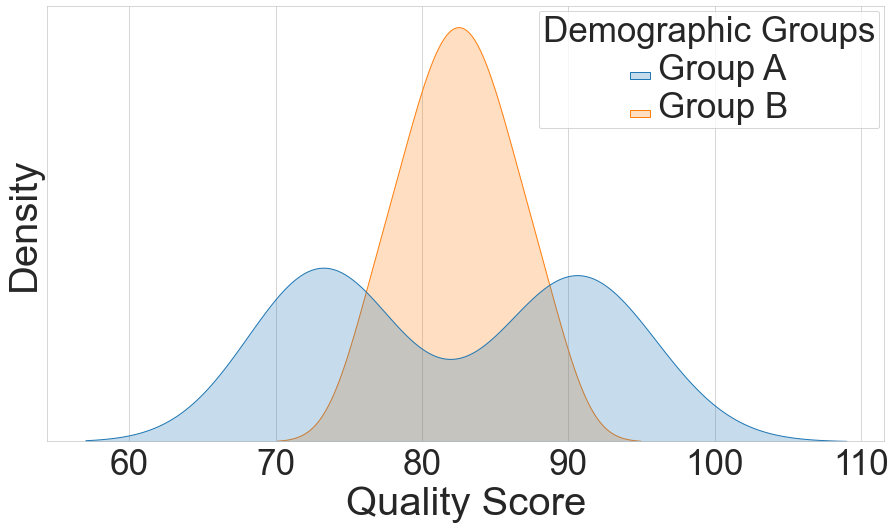}
    \caption{Fictitious quality component \textit{$Q_3$}: KDE Plot of the demographic score distribution} \label{fig3}
\end{minipage}\hfill
\begin{minipage}{0.45\textwidth}
    \centering
    \captionof{table}{Fictitious quality component \textit{$Q_3$}: Synthetic Mean and Median Quality Scores of different demographic groups}\label{tab5}
    \begin{tabular}{|l|c|c|c|}
    \hline
    & Group A & Group B \\
    \hline
    Mean & 81.95 & 82.5 \\
    \hline
    Median & 81.5 & 82.5 \\
    \hline
    LWM & 75.4 & 81.4 \\
    \hline
    \end{tabular}
\end{minipage}
\end{figure}

A fictitious scenario for demonstration the approach of LWM is visualised in Figure \ref{fig3}. Even though the quality scores of Group A and Group B have similar mean and median values (see Table \ref{tab5}), the underlying quality score distributions are very different. In this setup, more biometric samples of Group A would be rejected using lower operational thresholds than for Group B. However, as the LWM approach takes this behaviour into account, this results in a lower SQFR score\footnote{SQFR Scores in Table \ref{tab6} have been rounded to 3 decimal places as potential misinterpretations could occur with this setup}, as demonstrated in Table \ref{tab6}. 

The LWM approach achieves the lowest SQFR score in this scenario. When using the CSQFR mentioned in Section \ref{sec:cgc} with the LWM approach, the resulting LWM-GC-CSQFR becomes even lower, considering the different underlying distributions and placing greater emphasis on biased scenarios. On the other hand, one should be aware that the quality scores of Group A shown in Figure \ref{fig3} represents a rather unrealistic distribution in an operational environment and thus the advantage of the LWM approach could potentially be better in theory than in reality.

\begin{table}[htb]
\centering
\caption{SQFR results for quality component \textit{$Q_3$}}
\begin{tabular}{|c|c|}
\hline
SQFR & Quality component \textit{$Q_3$} \\
\hline
Mean-GC-SQFR & 0.997 \\
\hline
Median-GC-SQFR & 0.994 \\
\hline
LWM-GC-SQFR & 0.962 \\
\hline
LWM-GC-CSQFR & 0.889 \\
\hline
\end{tabular}
\vspace{0.5cm}
\label{tab6}
\end{table}

\subsection{Mean-Discard-Gap}
\label{mdpg}

The last DPM that we propose in this research paper is the Mean-Discard-Gap (MDG). This approach first calculates the proportion of the biometric samples of a demographic group that are below a certain number of relevant thresholds. Relevant thresholds are selected as follows:

\begin{equation}
    \text{Thresholds} = \{ \min(QS) + 1, \min(QS) + 2, \ldots, \max(QS) \}
\end{equation}
where \textit{QS} represents the set of quality scores from all demographic groups to be evaluated. Thus, thresholds are limited to quality scores that exist in the demographic data set. To avoid a zero-distance discard, the first relevant threshold starts at $\min(QS) + 1$. For all defined relevant threshold, a discard-percentage-gap is computed, which is the result of the distance between the minimum and maximum of the discard-percentage values across the groups. The final fairness measure is then derived by taking the mean discard-percentage-gap value of all min-max distances for all considered thresholds.

The adapted SQFR when using the MDG, resulting in the MDG-SQFR is calculated as follows:
\begin{equation}
 \text{MDG-SQFR} = 1 - \text{MDG}
\end{equation}

\begin{figure}[htb]
\centering
\begin{minipage}{0.45\textwidth}
    \centering
    \includegraphics[width=\textwidth]{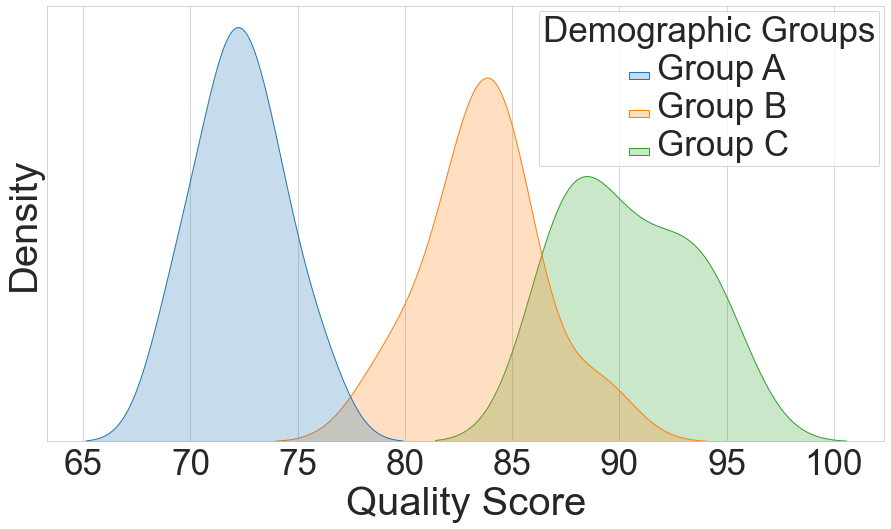}
    \caption{Fictitious quality component \textit{$Q_5$}: KDE Plot of the demographic score distribution} \label{fig5}
\end{minipage}\hfill
\begin{minipage}{0.45\textwidth}
    \centering
    \captionof{table}{Fictitious quality component \textit{$Q_5$}: Synthetic Mean and Median Quality Scores of different demographic groups}\label{tab9}
    \begin{tabular}{|l|c|c|c|c|c|}
    \hline
    & Group A & Group B & Group C  \\
    \hline
    Mean & 72.3 & 83.7 & 90.4 \\
    \hline
    Median & 72 & 83.5 & 90  \\
    \hline
    \end{tabular}
\end{minipage}
\end{figure}

A fictitious scenario for illustrating the behaviour of the MDG-SQFR can be seen in Figure \ref{fig5}.

On average, the quality scores of the three demographic groups differ from each other by 7-8 quality score points, which can be seen in Table \ref{tab9}. The resulting SQFR Scores with all proposed measures for the fictitious quality component \textit{$Q_5$} can be seen in Table \ref{tab10}.

\begin{table}[htb]
\centering
\caption{SQFR results for quality component \textit{$Q_5$}}
\begin{tabular}{|c|c|}
\hline
SQFR & Quality component \textit{$Q_5$} \\
\hline
Mean-GC-SQFR & 0.93 \\
\hline
Median-GC-SQFR & 0.93 \\
\hline
LWM-GC-SQFR & 0.93 \\
\hline
LWM-CSQFR & 0.81 \\
\hline
MDG-SQFR & 0.3 \\
\hline
\end{tabular}
\vspace{0.5cm}
\label{tab10}
\end{table}

Looking at the SQFR scores from Table \ref{tab10}, a clear trend can be seen: The resulting fairness score for the MDG-SQFR measure (0.3) is significantly lower than the previously presented measures (fairness scores of 0.93 and 0.81). This is due to the property of MDG that it considers the largest possible fairness difference per threshold and ignores groups in between. For this scenario, therefore, only groups A and C are considered for the fairness evaluation, when using MDG.

\section{Conclusion}
In this paper, several measures for the fairness assessment of biometric quality have been presented. A distinction can be made between DPMs that use the Gini coefficient or variations of it, and measures that treat fairness differently, such as the MDG-SQFR measure. 

In contrast to the GC-based measures, MDG-SQFR only considers the worst possible fairness difference per threshold $(max(discard\%) - min(discard\%))$ across demographic groups, not considering groups in between. Consequently, the number of demographic groups to be evaluated does not affect the resulting fairness score. A significant deviation of a single group (a disadvantaged group) is sufficient to receive a relatively low fairness score. 
This behavior can have an advantage over the GC-based measures for biased scenarios, as the resulting fairness score is generally lower. At the same time, however, this can also lead to a fairness score that is too low for scenarios where the quality score histograms are relatively similar across demographic groups, which is demonstrated in the scenario \textit{All groups receive similar QS} in Table \ref{tab11}.

The GC-based measures, on the other hand, are group size sensitive, consequently returning different fairness scores for different group sizes. However, it should be noted that the number of groups is usually rather limited (e.g., there are typically 2 groups for gender comparisons and ethnic and age-specific characteristics are often binned). Furthermore, the GC-based measures and variations may be less robust, as they use scalar approximation values such as the mean quality score of a demographic group, which may not accurately reflect the underlying quality score distribution. For a comparison and overview of different scenarios of all the measures presented in this paper, see Table \ref{tab11}. 

In general, if there is a preference of achieving even lower fairness scores for biased scenarios while slightly reducing the score of fairer scenarios, we recommend using a variation of the CSQFR over a variation of the SQFR. A promising CSQFR variant could be the LWM-GC-CSQFR, as it behaves similarly to the Mean-GC-CSQFR and additionally has the property of giving higher weight to lower quality scores. On the other hand, this weighting of the LWM-GC-CSQFR may not be necessary for quality score distributions in the field, as these are unlikely to include edge cases as demonstrated in Figure \ref{fig3} and therefore the simpler Mean-GC-CSQFR may already be sufficient for quality score distributions in the field.

Future work could focus on the application of SQFR on operational quality
assessment algorithms, including captured data in the field such as
passport enrolment images, kiosk enrolment images and border control
probe images. It will be of interest to analyse underlying variation of the
presented fairness rates over the full range of possible values.

\begin{table}[htb]
\centering
\caption{Comparison of proposed SQFR scores for different scenarios of 5 groups}
\vspace{0.5cm}
\resizebox{1\textwidth}{!}{
\setlength{\tabcolsep}{6pt}
\begin{tabular}{|l|c|c|c|c|c|c|c|c|c|c|c|c|}
\hline
QS scenarios of groups & \begin{tabular}[c]{@{}c@{}}\rotatebox{90}{Mean QS Group A}\end{tabular} & \begin{tabular}[c]{@{}c@{}}\rotatebox{90}{Mean QS Group B}\end{tabular} & \begin{tabular}[c]{@{}c@{}}\rotatebox{90}{Mean QS Group C}\end{tabular} & \begin{tabular}[c]{@{}c@{}}\rotatebox{90}{Mean QS Group D}\end{tabular} & \begin{tabular}[c]{@{}c@{}}\rotatebox{90}{Mean QS Group E}\end{tabular} & \begin{tabular}[c]{@{}c@{}}\rotatebox{90}{Mean-GC-SQFR}\end{tabular} & \begin{tabular}[c]{@{}c@{}}\rotatebox{90}{Mean-GC-CSQFR}\end{tabular} & \begin{tabular}[c]{@{}c@{}}\rotatebox{90}{LWM-GC-SQFR}\end{tabular} & \begin{tabular}[c]{@{}c@{}}\rotatebox{90}{LWM-GC-CSQFR}\end{tabular} & \begin{tabular}[c]{@{}c@{}}\rotatebox{90}{MDG-SQFR}\end{tabular} \\ \hline
One has strong bias& 31.4& 84.4& 84.9& 85.2& 86.8& 0.85& 0.61& 0.85& 0.62& 0.13 \\ \hline
Two have strong bias& 31.1& 26.7& 85& 85.1& 87.1& 0.72& 0.38& 0.73& 0.38& 0.12 \\ \hline
One has slight bias& 79.1& 85.6& 85& 85.1& 86.9& 0.98& 0.94& 0.98& 0.94& 0.52 \\ \hline
Two have slight bias& 76& 77.5& 85.6& 86.9& 85.8& 0.96& 0.89& 0.97& 0.9& 0.47 \\ \hline
All have similar QSs& 85.7& 87.5& 85.6& 86.6& 86.5& 0.99& 0.98& 0.99& 0.98& 0.71 \\ \hline
All have equal QSs& 87.5& 87.5& 87.5& 87.5& 87.5& 1& 1& 1& 1& 1 \\ \hline
All have different QSs& 87.5& 72.2& 25& 14.3& 47.3& 0.61& 0.22& 0.61& 0.22& 0.06 \\ \hline
\end{tabular}
}
\label{tab11}
\end{table}

\newpage
\bibliographystyle{splncs04}
\bibliography{bibliograph}

\end{document}